\documentclass{bvm} 

\makeatletter
\let\blx@rerun@biber\relax
\makeatother

\begin{document}

\newcommand{\bvmyear}{2024}

\selectlanguage{english} 

\title{Exploring Epipolar Consistency Conditions for Rigid Motion Compensation in In-vivo X-ray Microscopy}


\titlerunning{BVM \bvmyear}

\author{
	Mareike \lname{Thies} \inst{1}, 
	Fabian \lname{Wagner} \inst{1}, 
    Mingxuan \lname{Gu} \inst{1}, 
    Siyuan \lname{Mei} \inst{1},
    Yixing \lname{Huang} \inst{1}, 
    Sabrina \lname{Pechmann} \inst{2}, 
    Oliver \lname{Aust} \inst{3}, 
    Daniela \lname{Weidner} \inst{3}, 
    Georgiana \lname{Neag} \inst{3}, 
    Stefan \lname{Uderhardt} \inst{3}, 
    Georg \lname{Schett} \inst{3}, 
    Silke \lname{Christiansen} \inst{2,4}, 
    Andreas \lname{Maier} \inst{1}
}

\authorrunning{Thies et al.}

\institute{
\inst{1} Pattern Recognition Lab, FAU Erlangen-N\"urnberg\\
\inst{2} Fraunhofer Institute for Ceramic Technologies and Systems IKTS, Forchheim\\
\inst{3} Department of Rheumatology and Immunology, FAU Erlangen-N\"urnberg\\
\inst{4} Physics Department, Freie Universit\"at Berlin\\
}

\email{mareike.thies@fau.de}

\maketitle

\begin{abstract}
Intravital X-ray microscopy (XRM) in preclinical mouse models is of vital importance for the identification of microscopic structural pathological changes in the bone which are characteristic of osteoporosis. The complexity of this method stems from the requirement for high-quality 3D reconstructions of the murine bones. However, respiratory motion and muscle relaxation lead to inconsistencies in the projection data which result in artifacts in uncompensated reconstructions. Motion compensation using epipolar consistency conditions (ECC) has previously shown good performance in clinical CT settings. Here, we explore whether such algorithms are suitable for correcting motion-corrupted XRM data. Different rigid motion patterns are simulated and the quality of the motion-compensated reconstructions is assessed. The method is able to restore microscopic features for out-of-plane motion, but artifacts remain for more realistic motion patterns including all six degrees of freedom of rigid motion. Therefore, ECC is valuable for the initial alignment of the projection data followed by further fine-tuning of motion parameters using a reconstruction-based method. 
\end{abstract}

\section{Introduction}

Computed tomography (CT) imaging aids the reconstruction of internal structures of an object or organism by measuring X-ray projection images from multiple angles. When applied to living animals or humans, stillness of the subject is required throughout the scan as reconstruction algorithms rely on a highly accurate knowledge of the geometry between object, source, and detector pixels. This assumption cannot always be fulfilled due to inevitable motion in living organisms including cardio-respiratory motion or muscle relaxation. 

The X-ray microscope (XRM) that was utilized in this study is a specialized cone-beam CT scanner with a resolution down to $500\ts \rm nm$. Thus far, such scanners have mainly been used to investigate static samples in materials science, geoscience, or life science \cite{Jacobsen2019}. Recently, X-ray microscopy has garnered interest from the biomedical research field. The scope of our work is to improve the understanding of bone remodelling diseases like osteoporosis, at a high-resolution microscopic level, with the aid of XRM imaging of living mice. Whilst it is widely accepted that osteoporosis not only affects macroscopic parameters such as bone mass but also microscopic bone geometry, the dynamics of these changes from disease onset until diagnosis remain unclear to date \cite{Mader2013}. One central reason for this is the lack of longitudinal studies which jointly observe macroscopic and microscopic changes of bone structures in the same animal. With its non-invasive measurement procedure, high contrast between bone and soft tissue, and excellent resolution, XRM constitutes the ideal technology to enable such intravital longitudinal studies \cite{Wagner2022}. However, this requires imaging living animals and, consequently, the problem of subject motion has to be dealt with. 

As demonstrated in \cite{Mill2018}, motion from the mice's respiration and muscle relaxation can severely degrade the reconstructed XRM images up to a point where they become unusable for the attempted downstream tasks. There exists a plethora of works targeting rigid motion compensation in clinical settings \cite{Aichert2015,Frysch2015,Sisniega2017,Thies2023}. However, it is unclear how these algorithms translate to the special case of in-vivo XRM. As a first step, in this study, we investigate the applicability of epipolar consistency conditions (ECC) for motion compensation in XRM measurements of living mice. ECC are well-studied for artifact reduction in clinical cone-beam CT and have proven of great use for motion compensation in interventional C-arm head CT data \cite{Frysch2015,Preuhs2019} or empirical scatter and over-exposure correction \cite{Hoffmann2018,Preuhs2015}. Compared to such clinical use cases, the described XRM setting differs in a multitude of aspects, such as the scanner hardware, imaging protocols, geometry settings, and the structural properties of the object being scanned. We investigate the potential and limits of motion compensation with ECC for the given use case to assess whether it can pave the way toward high resolution in-vivo measurements for preclinical investigations of osteoporosis in a mouse model.   

\section{Materials and Methods}
\subsection{Epipolar Consistency Conditions}
The key idea of ECC for motion compensation is to exploit the redundancy in the projection data of a scan. As all acquired detector signals are projections of the same object under different geometries, they share common information which is related by the geometry settings. Motion introduces a mismatch in geometry and hence reduces the consistency in the data. Minimizing the inconsistency with respect to the geometry is a way to recover the mismatch introduced by subject motion. Aichert et al. \cite{Aichert2015} formally introduced ECC for transmissive radiation like X-rays and a way to apply them to cone-beam CT systems. The inconsistency between two corresponding epipolar lines $\mathbf{l}_0$ and $\mathbf{l}_1$ in the cone-beam case is computed via
\begin{equation}
    \Delta M = \mathrm{\frac{d}{dt}}\rho_{I_0}(\mathbf{l}_0) - \mathrm{\frac{d}{dt}}\rho_{I_1}(\mathbf{l}_1) \enspace ,
\end{equation}
where $\mathrm{\frac{d}{dt}}\rho_{I}$ refers to the derivative of the 2D Radon transform of projection image $I$ in $t$ direction, i.e., line distance to the image origin. In the ideal, motion-free case, this inconsistency is minimal for all pairs of epipolar lines. Hence, the motion compensation target function is the sum of all pairwise inconsistencies $\Delta M$ over all projection images in a scan and over all corresponding epipolar lines per pair. We refer the reader to the work of Aichert et al. \cite{Aichert2015} for a more in-depth explanation of ECC. The ECC target function is implemented based on the openly available source code by Aichert et al. (\url{https://github.com/aaichert/EpipolarConsistency}), but we develop additional Python bindings for the original code base written in C++. Since Python has evolved to one of the most common coding languages for research in medical image processing, we make our bindings available to the community as a prebuilt, ready-to-use Python package at \url{https://github.com/mareikethies/epipolar_consistency_python}. 

\subsection{Motion Simulation}
\begin{figure}[b]
    \centering
    \includegraphics[width=\textwidth]{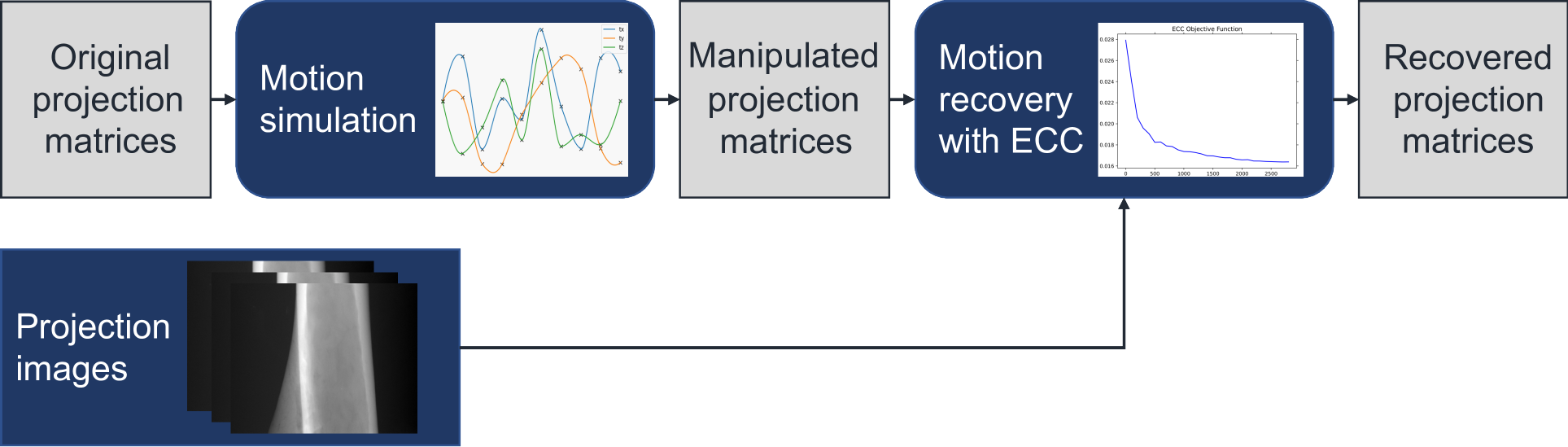}
    \caption{The original projection matrices are first manipulated and then recovered by minimizing the inconsistency in the projection images measured by ECC.}
    \label{3418-fig:pipeline}
\end{figure}

As XRM data of living mice are not currently available, we resorted to XRM measurements of an ex-vivo murine tibia sample. The scan has $N=1601$ projection images acquired on a short scan trajectory which we downsampled by a factor of $4$ for shorter run-time resulting in a size of $409 \times 509$ pixels with a pixel size of $5.6\ts \rm \text{\textmu} m$. Motion was artificially simulated by multiplying rigid transformation matrices  $\textrm{T}_i$ to the projection matrices $\textrm{P}_i$ with $i=1,\dots, N$. The idea is to find rigid transformations $\textrm{T}^*_i$ which recover the original, correct projection matrices from the manipulated ones (Fig.~\ref{3418-fig:pipeline}):
\begin{equation}
    \textrm{P}_i^{\textrm{recovered}} = \textrm{P}_i^{\textrm{motion}} \textrm{T}^*_i = \textrm{P}_i^{\textrm{original}} \textrm{T}_i \textrm{T}^*_i \enspace .
\end{equation}

Each of the rigid transformations has six degrees of freedom (3D rotation ($r_x, r_y, r_z$) and 3D translation ($t_x, t_y, t_z$)). Assuming that the motion of all projection images is independent, the problem has $6N$ degrees of freedom. Similar to prior work \cite{Preuhs2019}, we instead use a spline-based representation to enforce smoothness of the motion throughout the scan and reduce the dimensionality during optimization. A separate Akima-spline for each of the six motion parameters fits a smooth curve through specified nodes and reduces the number of free parameters from $6N$ to $6M$ where $M<N$ is the number of nodes for the spline. Motion simulation is performed using splines with $M=9$ nodes and a maximum perturbation of $\pm 50\ts \rm \text{\textmu} m$ and $\pm1^{\circ}$ for translation and rotation.      

\subsection{Optimization Details}
We minimize the projection-wise inconsistency quantified by ECC subject to the node values of the spline representations for the rigid transformation parameters. Three different scenarios are investigated relative to the plane spanned by the source positions during the scan: Optimization of (1) only out-of-plane parameters ($t_z, r_x, r_y$), (2) only in-plane parameters ($t_x, t_y, r_z$), and (3) all six parameters of rigid transformations. As suggested in the seminal work \cite{Aichert2015}, gradient-free optimization is performed using an adaptive Nelder-Mead downhill simplex optimizer \cite{Gao2012} for a maximum of $1000$ iterations for scenario (1) and (2) and $2000$ in scenario (3). All free parameters are initialized as zero and bounds are provided according to the maximum motion added during simulation. To assess the ability of the algorithm to recover the introduced motion independent of the modelling capacity of the spline, motion recovery is performed with a spline of the same type and with identical node positions as during motion simulation.  

\section{Results}
\begin{figure}[b]
  \begin{minipage}{0.24\textwidth}
    \hfill
  \end{minipage}\hfill
  \begin{minipage}{0.24\textwidth}
    \centering
    \includegraphics[width=\linewidth]{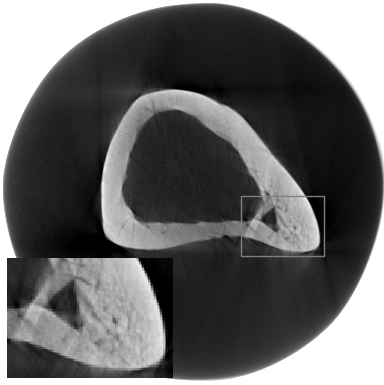}
    \subcaption{OOP only before optimization}
  \end{minipage}\hfill
  \begin{minipage}{0.24\textwidth}
    \centering
    \includegraphics[width=\linewidth]{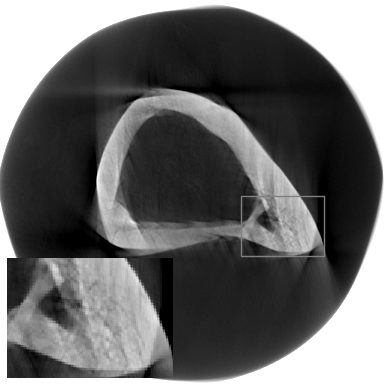}
    \subcaption{IP only before optimization}
  \end{minipage}\hfill
  \begin{minipage}{0.24\textwidth}
    \centering
    \includegraphics[width=\linewidth]{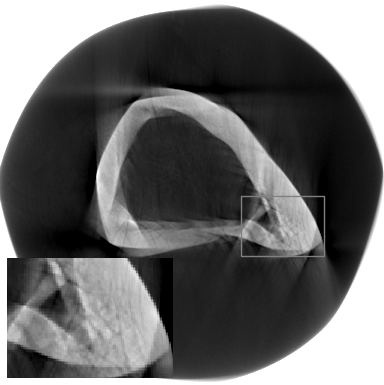}
    \subcaption{Full rigid before optimization}
  \end{minipage}

  \begin{minipage}{0.24\textwidth}
    \centering
    \includegraphics[width=\linewidth]{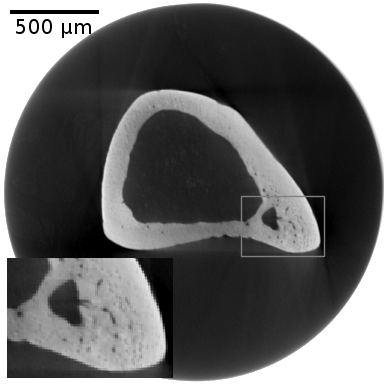}
    \subcaption{Original}
  \end{minipage}\hfill
  \begin{minipage}{0.24\textwidth}
    \centering
    \includegraphics[width=\linewidth]{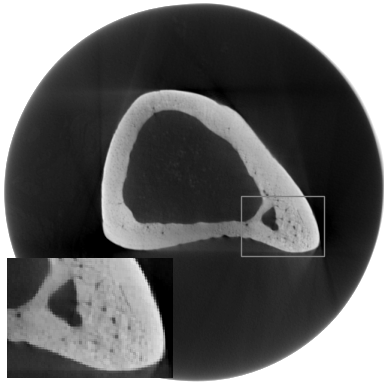}
    \subcaption{OOP only after optimization}
    \label{3418-fig:oop_after}
  \end{minipage}\hfill
  \begin{minipage}{0.24\textwidth}
    \centering
    \includegraphics[width=\linewidth]{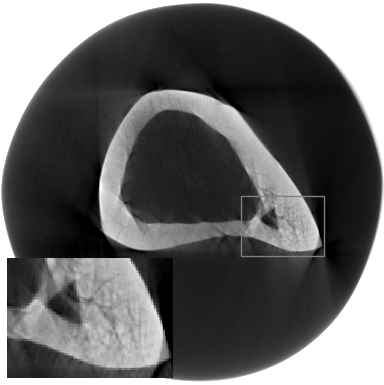}
    \subcaption{IP only after optimization}
  \end{minipage}\hfill
  \begin{minipage}{0.24\textwidth}
    \centering
    \includegraphics[width=\linewidth]{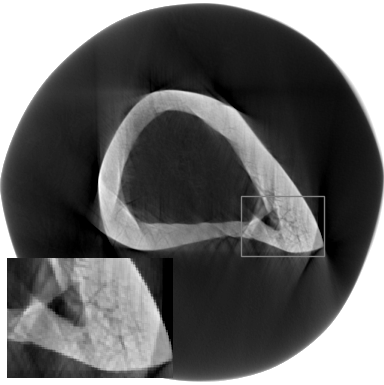}
    \subcaption{Full rigid after optimization}
  \end{minipage}

	\caption{Reconstruction results before (upper row) and after motion compensation (lower row). We show an off-center slice because effects of rotation do not exhibit themselves well in the center slice. (d) is the ground truth reconstruction, (a) and (e) correspond to optimization of out-of-plane (OOP) parameters only, (b) and (f) correspond to optimization of in-plane (IP) parameters only, and (c) and (g) correspond to a full rigid optimization. }
	\label{3418-fig:results}
\end{figure} 
Reconstructions are performed by an FDK algorithm \cite{Thies2022} using the downsampled projection images and the original, manipulated or recovered projection matrices. Fig.~\ref{3418-fig:results} shows reconstruction results of the three investigated scenarios. The internal structure of the bone can only be recovered if just out-of-plane motion is considered (Fig.~\ref{3418-fig:oop_after}). For the other two cases with in-plane motion only and full rigid transformations, artifacts remain. Nevertheless, the general shape of the bone is improved. These findings are also represented quantitatively in Tab.~\ref{3418-tab:results}. Mean-squared error (MSE) and structural similarity (SSIM) are improved in each case, but very high agreement with the ground truth reconstruction can only be achieved in the simplified setup of just out-of-plane motion. The error of transformation parameters is reduced in all cases but the reduction is only marginal for rotations in $z$.

\begin{table}[t]
\caption{The error in reconstruction domain is measured by MSE and SSIM to the ground truth reconstruction. The L$_1$ error quantifies the mean absolute error over all projection images for the considered transformation parameters. All values are presented as \textit{before $\rightarrow$ after optimization}.}
\label{3418-tab:results}
\centering
\resizebox{0.9\textwidth}{!}{%
\begin{tabular*}{\textwidth}{l@{\extracolsep\fill}rrr}
\hline
 & \emph{out-of-plane only} & \emph{in-plane only} & \emph{rigid complete}\\ \hline
\# parameters & 27 & 27 & 54\\
MSE [$*10^{-8}$] & 0.59 $\rightarrow$ 0.17 & 3.53 $\rightarrow$ 1.49 & 3.64 $\rightarrow$ 1.93 \\
SSIM & 0.91 $\rightarrow$ 0.94 & 0.80 $\rightarrow$ 0.87 & 0.80 $\rightarrow$ 0.84 \\
L1 error t$_x$ [$\rm \text{\textmu} m$] & x & 20.95 $\rightarrow$ 12.37 & 20.95 $\rightarrow$ 15.23 \\
L1 error t$_y$ [$\rm \text{\textmu} m$] & x & 23.98 $\rightarrow$ 4.83 & 23.98 $\rightarrow$ 8.91 \\
L1 error t$_z$ [$\rm \text{\textmu} m$] & 21.19 $\rightarrow$ 12.67 & x & 21.19 $\rightarrow$ 7.92 \\
L1 error r$_x$ [deg] & 0.41 $\rightarrow$ 0.18 & x & 0.41 $\rightarrow$ 0.19 \\
L1 error r$_y$ [deg] & 0.33 $\rightarrow$ 0.10 & x & 0.33 $\rightarrow$ 0.19 \\
L1 error r$_z$ [deg] & x & 0.44 $\rightarrow$ 0.43 & 0.44 $\rightarrow$ 0.41 \\
 \hline
\end{tabular*}}
\end{table}

\section{Discussion}
The results imply that it is generally feasible to improve the reconstruction quality of motion-corrupted XRM scans using ECC-based optimization. The overall bone shape is improved in the motion compensated reconstructions and, for the case of just out-of-plane parameters, even the small structures inside the bone are recovered. For in-plane and full rigid motion, restoring the small-scale structures within the bone has shown to be difficult. It is apparent that the algorithm cannot compensate all six motion types with the same accuracy, especially rotation around $z$ is only poorly recovered. This can be explained as it occurs within the epipolar planes and, therefore, does not have a large influence on the target function. Also, the considered bone is highly symmetric regarding rotation around its long axis which further complicates the formulation of a discriminative target function. Additionally, subject motion is not the only factor influencing the ECC target function. Other physical phenomena such as noise, scatter or beam hardening can also introduce mismatch between projection images and disturb the motion correction when close to the optimal solution. We further acknowledge that by using identical node positions between motion simulation and recovery, we do not mimic a realistic setup. But as the algorithm does not yield satisfactory results for all parameters in this simplified scenario, we did not include results of more complicated settings. Further, the presented algorithm will experience major performance drops as soon as the projections are laterally truncated, i.e., the object does not fully fit into the field-of-view. The ECC target function compares plane integrals through the measured object evaluated in different projection images. If projection images are truncated, the integral can only be evaluated in the non-truncated region which introduces an error and makes the ECC target function hardly usable for motion compensation. When imaging at smaller pixel sizes or larger anatomical regions of the tibia such as its proximal end (knee region), truncation might occur due to the limited field-of-view of the XRM scanner. 

In conclusion, while ECC is valuable for a first alignment of motion corrupted data, reconstruction-based methods \cite{Sisniega2017} seem to be more suitable for motion compensation in in-vivo XRM studies because they do not depend on non-truncated projection images and are more sensitive to changes in all six parameters of rigid motion.

\begin{acknowledgement}
	The research leading to these results has received funding from the European Research Council (ERC) under the European Union’s Horizon 2020 research and innovation program (ERC Grant No. 810316). The authors gratefully acknowledge the HPC resources provided by NHR@FAU. The hardware is funded by the German Research Foundation (DFG).
\end{acknowledgement}

\printbibliography

\end{document}